\documentclass[runningheads]{llncs}
\usepackage[T1]{fontenc}
\usepackage{amsmath} 
\usepackage{graphicx}
\usepackage{color}
\usepackage{xcolor} 
\usepackage{hyperref} 
\usepackage{url} 

\usepackage{marvosym}
\usepackage{hyperref}  
\usepackage{multirow}  
\begin{document}

%
\title{Exploring Gender Bias in Alzheimer’s Disease Detection: Insights from Mandarin and Greek Speech Perception\thanks{This work was supported by the National Social Science Foundation of China (Grant No.23AYY012)}}

\titlerunning{Gender Bias in AD Perception from Mandarin and Greek Speech}
%
\author{Liu He\inst{1} \and
Yuanchao Li\inst{2} \and
Rui Feng\inst{1}\and
XinRan Han \inst{3} \and
Yin-Long Liu \inst{1} \and 
Yuwei Yang \inst{4} \and 
Zude Zhu \inst{4} \and 
Jiahong Yuan \inst{1}\textsuperscript{\Letter}}

\authorrunning{He et al.}
%
\institute{
    University of Science and Technology of China, Hefei Anhui 230026, China \\
    \email{heliummn@mail.ustc.edu.cn, fengruimse@mail.ustc.edu.cn, lyl2001@mail.ustc.edu.cn, jiahongyuan@ustc.edu.cn}
    \and 
    %
    University of Edinburgh, Edinburgh EH8 9YL, UK \\
    \email{yuanchao.li@ed.ac.uk}
    \and 
    %
    University of Minnesota, Minneapolis MN 55455, USA \\
    \email{han00610@umn.edu}
    \and 
    %
    Jiangsu Normal University, Xuzhou Jiangsu 221116, China \\
    \email{18352261722@163.com, 6020150015@jsnu.edu.cn}
}

\maketitle             
\begin{abstract}
Gender bias has been widely observed in speech perception tasks, influenced by the fundamental voicing differences between genders. This study reveals a gender bias in the perception of Alzheimer’s Disease (AD) speech. In a perception experiment involving 16 Chinese listeners evaluating both Chinese and Greek speech, we identified that male speech was more frequently identified as AD, with this bias being particularly pronounced in Chinese speech. Acoustic analysis showed that shimmer values in male speech were significantly associated with AD perception, while speech portion exhibited a significant negative correlation with AD identification. Although language did not have a significant impact on AD perception, our findings underscore the critical role of gender bias in AD speech perception. This work highlights the necessity of addressing gender bias when developing AD detection models and calls for further research to validate model performance across different linguistic contexts.

\end{abstract}
\section{Introduction}

Gender is known to play a significant role in cognitive aging and the progression of Alzheimer’s Disease (AD) \cite{li2014sex,nebel2018understanding}. Research suggests that various biological and psychosocial factors contribute to differences in AD risk and progression across genders. For instance, some studies indicate that females may be at a higher risk of developing AD dementia due to differences in brain structure, responses to psychosocial stress, and hormonal influences \cite{zhu2021alzheimer}. Additionally, cognitive decline following the onset of Mild Cognitive Impairment (MCI) or AD dementia has been reported to progress more rapidly in some female populations \cite{irvine2012greater}.

From a neurocognitive perspective, differences in cognitive function have been observed, with verbal memory often being more advantageous in females. However, some studies suggest that these differences may also influence the trajectory of AD-related cognitive decline \cite{bayles1999gender}, highlighting the importance of considering gender as a factor in AD research.

Beyond biological influences, socio-cultural factors also shape AD perception. Variables such as educational attainment and professional opportunities have been linked to AD risk  \cite{lopez2003risk}. Historically, disparities in access to education and career advancement may have contributed to variations in AD risk across genders, particularly in underdeveloped regions. Additionally, depression, which has been identified as a risk factor for cognitive decline, is more frequently diagnosed in females. Studies suggest that moderate to severe depressive symptoms may have a significant impact on AD risk \cite{makizako2016comorbid,sundermann2017sex}.

Despite these research on gender differences in AD risk and progression, the influence of gender on speech-based AD perception remains largely unexplored. To address this gap, this study is the first to examine whether gender plays a role in how AD is perceived in speech. Specifically:
\begin{itemize}
    \item We investigate whether speech from a particular gender is more likely to be identified as AD through human perception experiments.
    \item We analyze how acoustic features such as shimmer, jitter, and speech portion contribute to these perceptions. The study is conducted across two linguistically diverse contexts, Mandarin Chinese and Greek, to enhance generalizability.
\end{itemize}

Our experimental results reveal that male speech is more frequently perceived as indicative of AD, particularly in Mandarin. Additionally, specific speech features, such as shimmer, significantly influence these perceptions. These findings underscore the need to account for potential gender biases in the development of speech-based AD detection models and highlight the importance of further research to ensure the generalizability and fairness of diagnostic tools.

\section{Related Work}

\subsection{Speech Features and Gender in AD}


The analysis of speech features has become a valuable approach for detecting AD.  A review of the literature reveals that past research has converged on several key acoustic domains to differentiate individuals with AD from Healthy Controls (HC). One primary focus is the temporal and fluency characteristics of speech. Features in this category, such as the overall speech portion (the ratio of speaking time) and the prevalence of long silences, have been shown to be effective indicators of cognitive decline\cite{yamada2022speech,dutta2024universal,pastoriza2022speech,de2018changes}. Another critical domain involves micro-acoustic perturbations in voice quality, which can reflect deteriorating neuromuscular control. In this area, features like Jitter (period-to-period variations in fundamental frequency) and Shimmer (period-to-period variations in amplitude) are frequently identified as significant markers \cite{yamada2022speech,meilan2014speech,XIU2025286}. Other studies have also pointed to changes in vocal energy dynamics, such as deviations in delta energy, as being particularly indicative of AD, especially in male speakers \cite{khodabakhsh2015evaluation}.

However, a significant challenge arises from the observation that many of these features appear to show less effectiveness in female speakers, suggesting that gender may be a critical modulating factor in the reliability of speech-based AD detection. Furthermore, research has shown that speech expressiveness varies between genders, with differences observed in aspects such as emotional prosody and intonation patterns \cite{li2019improved}. Male speakers with AD tend to display less expressiveness than female speakers, which may further affect speech-based AD detection \cite{khodabakhsh2015evaluation}.

\subsection{Perceptual Bias in Speech-Based AD Detection}

Listeners’ perceptions of speech are shaped by various cognitive and social factors, including gender stereotypes associated with vocal characteristics \cite{li2022robotic}. Research suggests that pitch, intonation, and articulation patterns influence how speaker traits are interpreted (e.g., certain genders are expected to exhibit higher pitch and stronger sibilance), often giving rise to perceptual biases \cite{campbell2011intersecting,levon2014categories}. Such biases may contribute to differences in how cognitive decline is recognized and assessed. For instance, speech difficulties or memory loss in male speakers may be more readily attributed to disease, as these symptoms conflict with prevailing societal expectations that associate male speech with rationality and stability. In contrast, similar symptoms in female speakers are more likely to be dismissed as emotional responses or part of the natural aging process \cite{lewin2001sex}. As a result, this perceptual bias may contribute to the underdiagnosis or misdiagnosis of cognitive decline in female individuals, as their symptoms are less likely to be interpreted as indicators of AD and more often regarded as age-related or emotional nature.

\section{Methods}

\subsection{Datasets}
In this study, we utilized both Greek speech data from the ADReSS-M challenge and Mandarin speech samples from the NCMMSC2021 challenge to explore the potential impact of gender, language, and speech features on the perception of AD and HC groups.

\subsubsection{Greek Data from ADReSS-M challenge}

The ADReSS-M challenge was hosted by the ICASSP 2023 conference\footnote{https://luzs.gitlab.io/madress-2023/}. Its dataset includes audio recordings of picture descriptions from 148 AD patients and 143 HCs, recorded in both English and Greek. The Greek recordings assess participants’ verbal fluency and mood through a picture description task. Participants were shown an image depicting a lion lying with a cub in the desert while eating (see Figure~\ref{fig:lion}) \cite{luz2024overview}. The dataset was carefully split, ensuring balanced representation of age and gender. We used the full set of Greek language data for our analysis.

\subsubsection{Mandarin Data from NCMMSC2021 challenge}

\begin{figure}[t]
  \centering
  \begin{minipage}{0.46\linewidth}
    \centering
    \includegraphics[width=\linewidth]{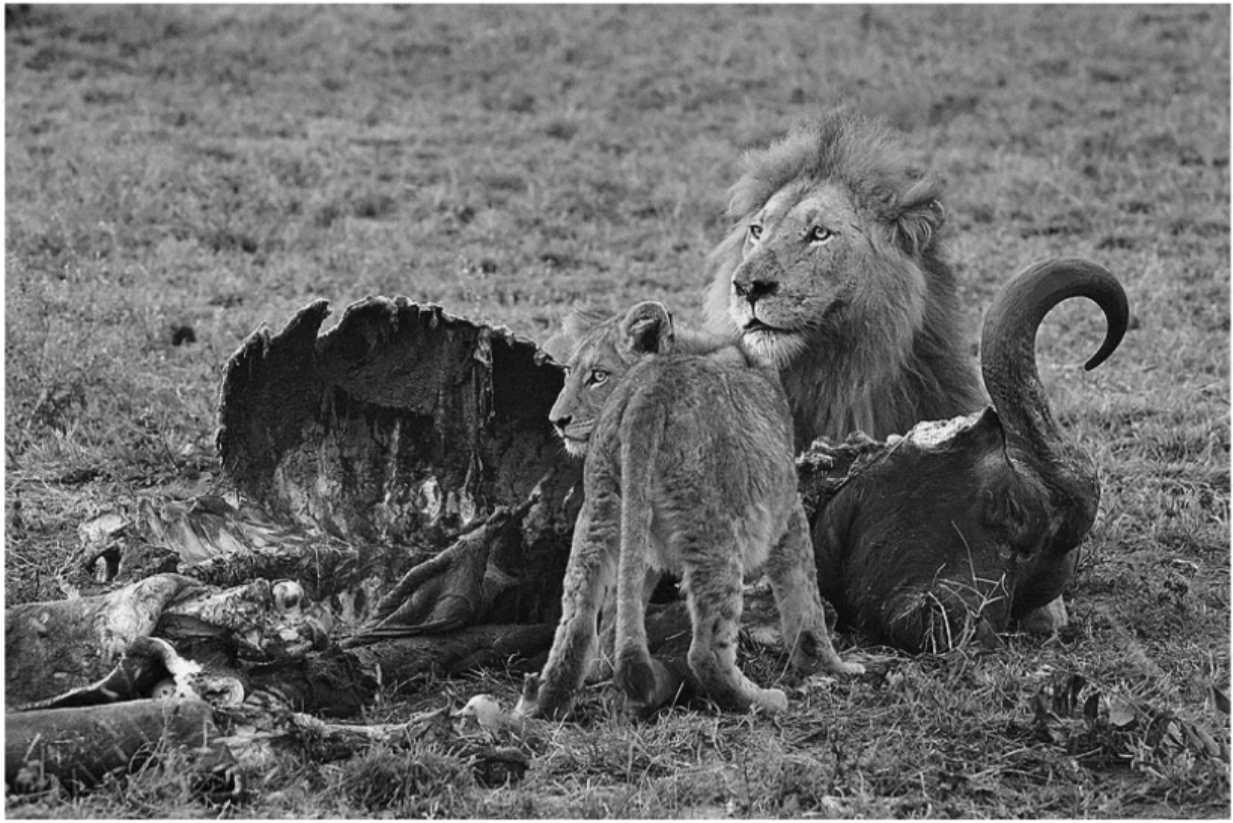}
    \caption{Lion picture for Greek AD.}
    \label{fig:lion}
  \end{minipage} \hfill
  \begin{minipage}{0.47\linewidth}
    \centering
    \includegraphics[width=\linewidth]{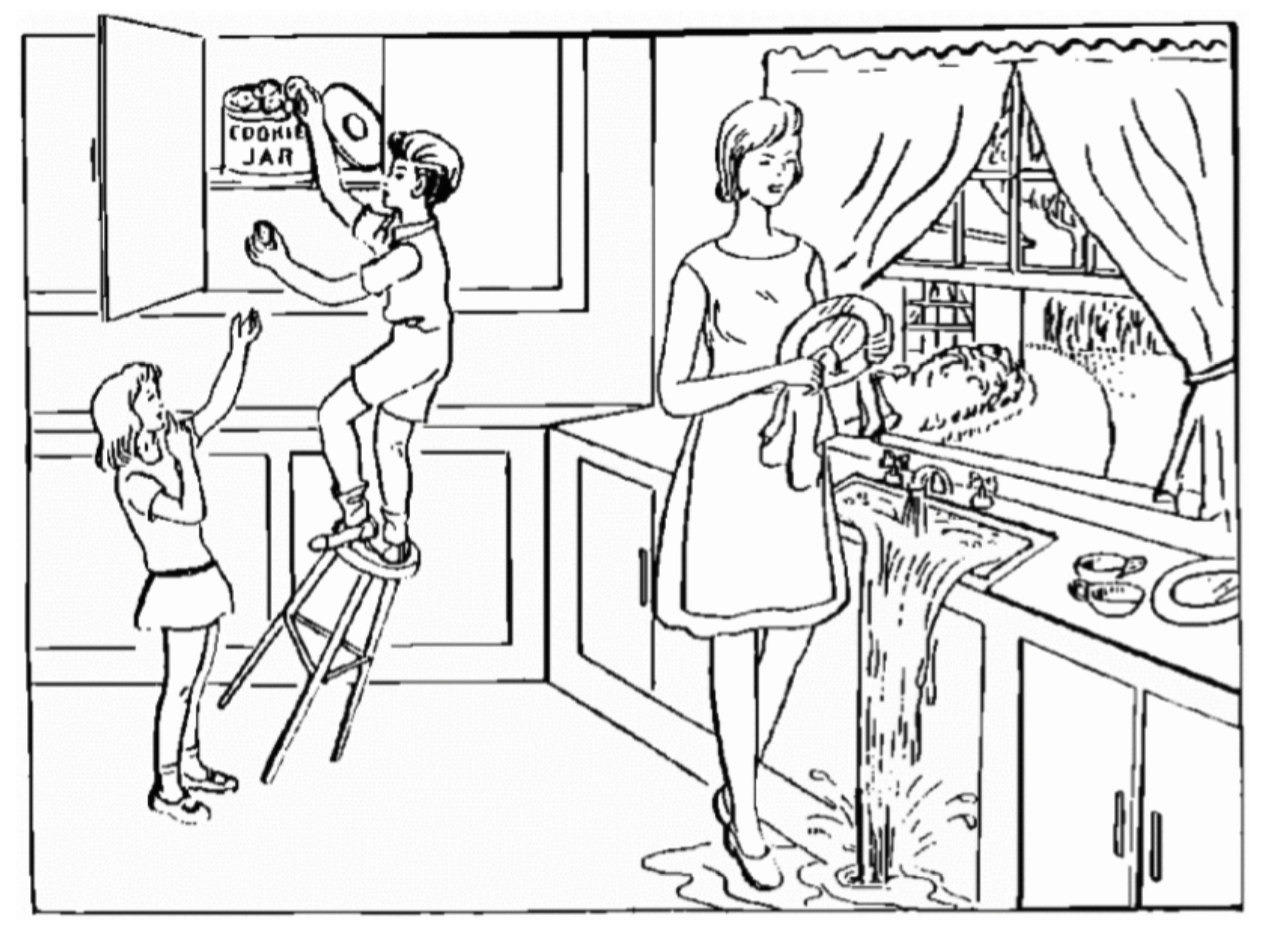}
    \caption{Cookie Theft picture for Mandarin AD.}
    \label{fig:cookie}
  \end{minipage}
\end{figure}

The Mandarin data for this study was sourced from the NCMMSC2021 AD Recognition Challenge, with data provided by anonymous University 1. The recordings involved 124 Chinese speakers, consisting of 26 AD patients, 44 HCs, and 54 MCI subjects. The Mandarin audio recordings also consist of picture descriptions, where participants were asked to describe the Cookie Theft picture (see Figure~\ref{fig:cookie}) from the Boston Diagnostic Aphasia Examination test \cite{becker1994natural}. All participants were native Mandarin speakers, and the description task was identical to that used in the ADReSS-M challenge.

\subsection{Experiment Design}
Stimuli were drawn from ADReSS-M challenge in Greek \cite{luz2024overview}, and from the NCMMSC2021 challenge in Mandarin. For each language, 30 picture description utterances (15 AD/15 HC) with durations ranging from 0 to 1 minute were chosen. Each utterance consisted of dialogues between healthcare professionals and participants. The audio characteristics of Greek and Chinese recordings were generally comparable. Sociodemographic attributes including gender, age, and education level were documented. Gender and age distributions were balanced as much as possible between the two language groups, though complete parity was unattainable due to limitations in the Greek corpus. The experiment utilized 30 audio recordings per language, with AD and HC samples randomly ordered within each language group.

Sixteen students (8 male, 8 female) from the anonymous University 2 participated, with a mean age of 23 years. All participants were native Mandarin speakers without formal Greek language training. The experiment employed a compensation scheme, with top performers receiving commendation.

The stimuli were presented through the webMUSHRA interface \cite{schoeffler2018webmushra}. For each language condition, participants first received background information about AD and HC characteristics. They then listened to audio clips and performed two tasks: 1) classifying each sample as AD or HC, 2) identifying specific time segments influencing their decisions through a timeline annotation interface. Participants additionally provided free-text rationales for each classification. The Greek perception experiment preceded the Mandarin condition to prevent linguistic interference.

\subsection{Data Process}

Three key audio features were extracted from 60 Mandarin Chinese and Greek audio samples: \textit{Speech Portion}, \textit{Jitter}, and \textit{Shimmer}. Each audio file was resampled to 16 kHz to standardize the sampling rate, and the audio was converted to a mono channel to ensure uniformity in the analysis.

\subsubsection{Speech Portion}

Speech activity detection was performed using the Silero Voice Activity Detector (VAD) model\footnote{https://github.com/snakers4/silero-vad}, which is based on deep learning techniques to accurately classify audio segments as speech or non-speech (e.g., silence, noise). The model was pre-trained and used for both datasets with consistent parameters to maintain comparability.
	
The detection threshold was set to 0.5, which represents the model’s sensitivity to speech signals. The minimum speech duration was set to 250 ms, and the minimum silence duration was set to 100 ms, ensuring that short speech segments and brief pauses were considered appropriately.

The ratio of total speech duration to total audio duration was calculated equation \ref{eq2}. This was done by summing the durations of all speech segments identified by the VAD model and dividing it by the total length of the audio recording. The result represents the proportion of the audio that is speech, excluding non-speech segments such as pauses or background noise. 

\begin{equation}
    \text{Speech Portion} = \frac{\sum_{i} (t_{\text{end}} - t_{\text{start}})}{\text{Total Audio Duration}}
\label{eq2}
\end{equation}

\subsubsection{Jitter and Shimmer}

Jitter and shimmer were computed through Parselmouth \cite{parselmouth}, a Python wrapper for Praat voice analysis toolkit. Jitter was measured using Praat's ``Get jitter (local)'' algorithm \cite{praat}, calculating the average relative difference between consecutive pitch periods. A time window of 0.02 seconds was used to assess the period-to-period variations in pitch, with a maximum allowed period difference of 1.3. Shimmer was obtained via ``Get shimmer (local)'' method, quantifying the mean relative amplitude variation between adjacent cycles. Shimmer was calculated using a time window of 0.02 seconds, with a maximum period difference of 1.3 and a maximum amplitude difference of 1.6.

\subsubsection{Perception Weighted Score}

 For each stimulus, the responses from 16 listeners were assigned binary values: $1$ if the listener classified the stimulus as AD and $0$ if the listener classified it as HC. The Perception Weighted Score (PWS) for each stimulus was then calculated by weighting the responses across all listeners. Since all listeners are assigned equal weight, the PWS for a stimulus $i$ is computed as the simple mean of the binary responses from all listeners, as shown by the following formula \ref{PWS}:
 
 \begin{equation}
         PWS_i = \frac{1}{N} \sum_{j=1}^{N} S_{ij} 
 \label{PWS}
 \end{equation}
 
 where \( S_{ij} \) denotes the response of listener \( j \) to stimulus \( i \), and \( N \) is the total number of listeners (i.e., 16 in this study).

This formula calculates the average classification of each stimulus across all listeners, providing an overall perception score based on the collective judgment.

\subsection{Statistical Analysis}

To analyze the influence of gender and speech features on the perception of AD versus HC, we employed a Generalized Linear Mixed-Effects Model (GLMER) \cite{bates2015fitting}. This method was chosen due to its ability to account for both fixed effects, such as speaker gender and speech characteristics (e.g., speech portion, jitter, shimmer), and random effects, such as variability among individual listeners. The GLMER approach is particularly well-suited for hierarchical or nested data structures, where data points (e.g., speech samples) are nested within higher-level groups (e.g., listeners). This makes it a more appropriate choice compared to simpler models like ANOVA, which do not account for the variability due to random effects \cite{matuschek2017balancing}. The GLMER model was formulated as:

\begin{equation}
\label{equation:model} 
\begin{split}
    \text{Response}_{ij} \sim \beta_0 &+ \beta_1 \text{Gender}_{ij} + \beta_2 \text{Language}_{ij} + \beta_3 \text{Speech\ Portion}_{ij} \\
    &+ \beta_4 \text{Jitter}_{ij} + \beta_5 \text{Shimmer}_{ij} + (1 | \text{Listener\ ID}_i) + \epsilon_{ij}
\end{split}
\end{equation}

where $\text{Response}_{ij}$ is the perception outcome (AD vs. HC) for the $j$-th speech sample from the $i$-th listener; $(1| \text{Listener\ ID}_i)$ represents random effect term for listener-specific variability; $\beta_0$ denotes intercept term; $\beta_1$-$\beta_5$ means fixed effect coefficients for $\text{Gender}$, $\text{Language}$, $\text{Speech\ Portion}$, $\text{Jitter}$, $\text{Shimmer}$;  and $\epsilon_{ij}$ for the residual error term.

This model was fit using maximum likelihood estimation, with Laplace approximation for the likelihood of the random effects \cite{bates2015fitting}. The Laplace approximation was chosen due to its efficiency in models with non-normal random effects distributions. Model selection was based on the Akaike Information Criterion and Bayesian Information Criterion, ensuring the model’s appropriateness for the dataset.

Such a generalized linear mixed-effects model allows for the inclusion of both fixed effects, such as speaker characteristics and speech features, and random effects, capturing individual listener variability in speech perception. By using GLMER, we ensure that our analysis accounts for the nested structure of the data (with listeners as random effects) while simultaneously testing the influence of gender and various speech features on the perception of AD and HC.

\section{Perception Experiment Results}

\subsection{Perception Distribution}

We analyzed the distribution of perception scores across language and gender using PWS, aggregating listeners’ responses.

\subsubsection{Language}

\begin{figure}[t]
  \centering
  \begin{minipage}{0.49\linewidth}
    \centering
    \includegraphics[width=\linewidth]{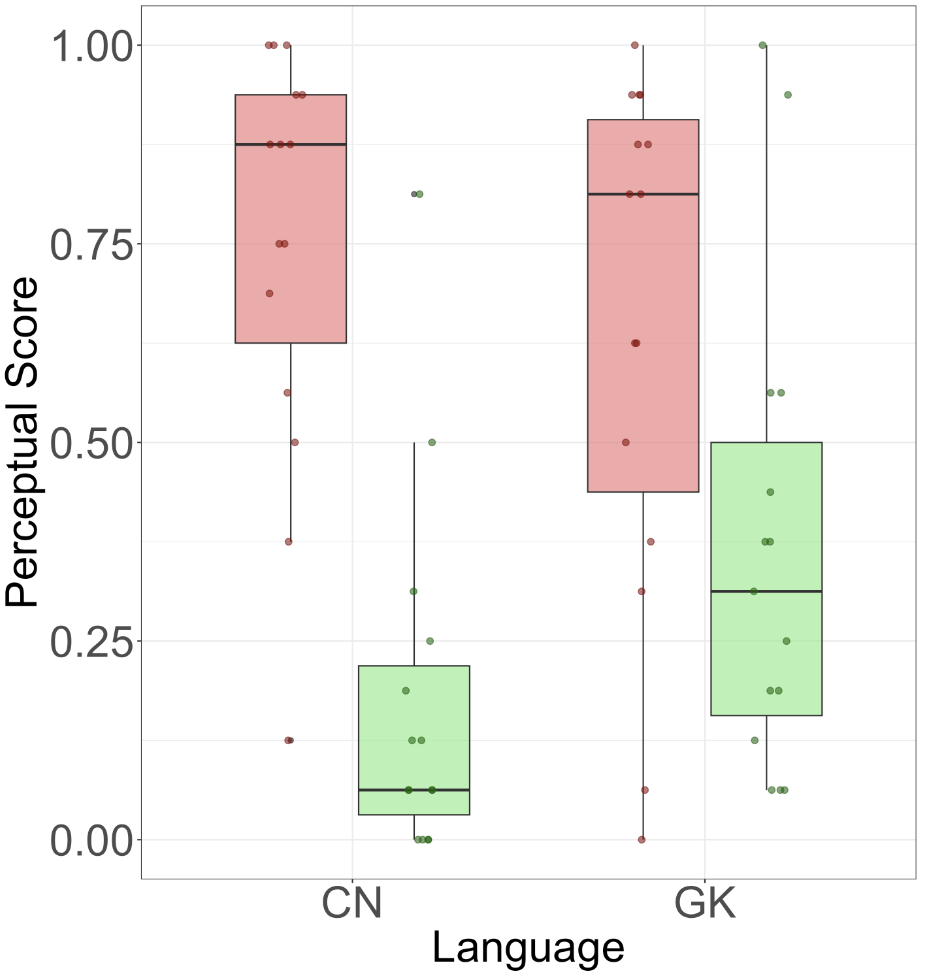}
    \caption{Perceptual Scores Distribution by Language.}
    \label{fig:language}
  \end{minipage} \hfill
  \begin{minipage}{0.49\linewidth}
    \centering
    \includegraphics[width=\linewidth]{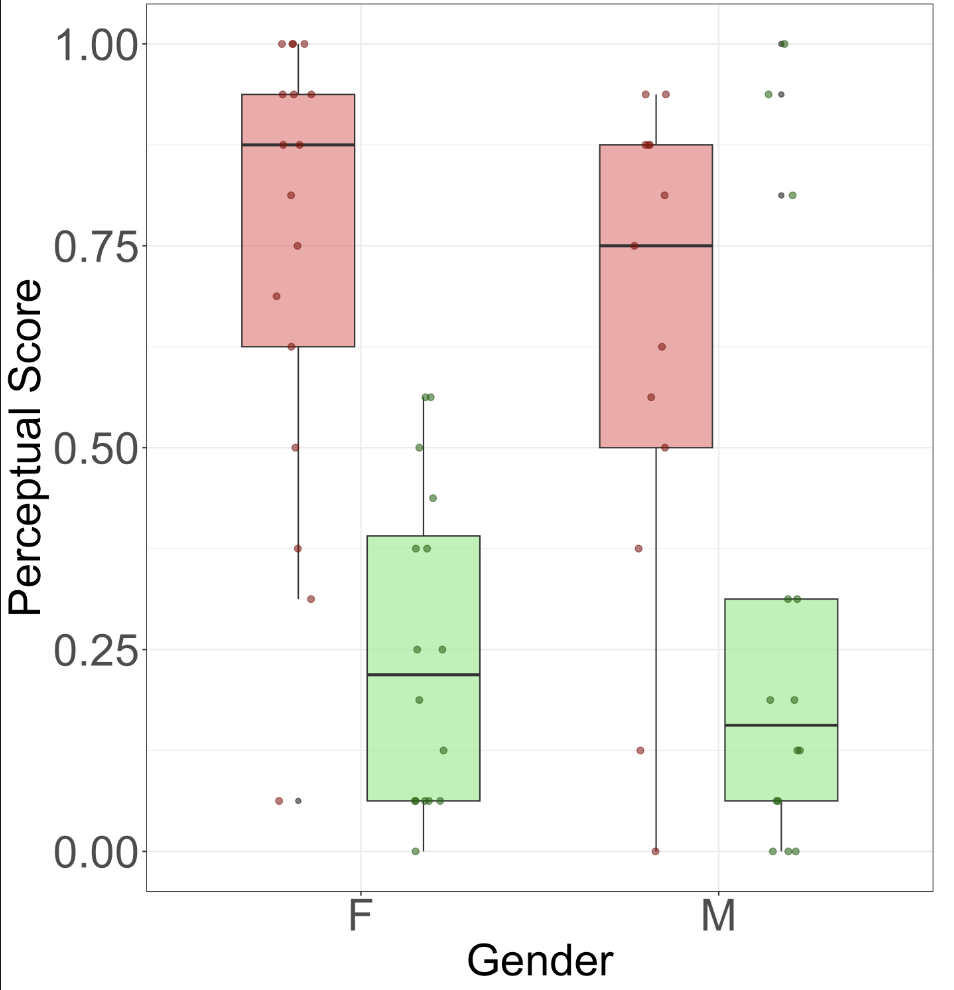}
    \caption{Perceptual Scores Distribution by Gender.}
    \label{fig:gender}
  \end{minipage}
\end{figure}

Figure~\ref{fig:language} shows the distribution of perceptual scores by language (Mandarin Chinese: CN, Greek: GK), with scores color-coded based on the true labels (AD or HC). In the CN dataset, AD stimulus consistently received higher and more concentrated perceptual scores, while HC stimulus were associated with lower scores. In contrast, the GK dataset exhibited considerable overlap in the perceptual scores between AD and HC stimulus, with HC stimulus demonstrating a narrower distribution and lower perceptual scores. These findings suggest a higher capacity for recognizing AD speech among Chinese listeners in their native language.

\subsubsection{Gender}

Figure~\ref{fig:gender} shows the distribution of perceptual scores by gender (M, F), with scores color-coded based on the true labels (AD or HC). For females, the AD stimulus consistently received higher and more concentrated perceptual scores, while HC stimuli were associated with lower scores. In contrast, some males demonstrated higher perceptual scores for HC stimuli. These findings suggest a gender bias in perceptual scores, with some males exhibiting higher scores for HC stimuli.


\subsection{Statistical Analysis: Validating Gender bias in AD Perception}

The results from the GLMER model (Equation~\ref{glmer1}) analyzing data from both Chinese listeners who listened CN and GK speech data, are summarized in Table~\ref{tab:vertical}. The analysis shows that gender significantly influences the perception of AD, with male speech being more likely to be identified as AD compared to female speech ($p = 0.035$, Odds Ratio = 1.41). In contrast, language did not significantly affect the recognition of AD or HC ($p = 0.930$), suggesting no notable impact of language on AD perception. Among speech features, the proportion of speech in the recording had a significant negative effect on AD perception ($p < 0.001$, Odds Ratio = 0.22), while shimmer also played a crucial role ($p = 0.020$, Odds Ratio = 0.64). However, jitter did not significantly impact AD perception ($p = 0.911$), indicating that it was not a distinguishing factor in this study.

\begin{equation}
\begin{split}
  \text{Response} &\sim   \text{Gender} + \text{Language} +  \text{Speech\ Portion}  \\ 
  &+ \text{Jitter} + \text{Shimmer} + (1 | \text{Listener\ ID})
  \label{glmer1}
\end{split}    
\end{equation}

\begin{table}[t!]
  \caption{All (CN\&GK N=960), CN (N=480), GK (N=480) Generalized Linear Mixed Model Results.}
  \label{tab:vertical}
  \centering
  \scalebox{0.95}{
  \begin{tabular}{|l|l|r|r|r|}
    \hline
    \textbf{} & \textbf{Pred} & \textbf{Coef(SE)} & \textbf{OR [95\% CI]} & \textbf{p-val} \\
    \hline
    \multirow{5}{*}{All} 
    & G(M) & \textbf{0.34 (0.16)} & \textbf{1.41 [1.03,1.94]} & $0.035^{*}$ \\
    & Speech Portion & \textbf{-1.53 (0.11)} & \textbf{0.22 [0.17,0.26]} & $<0.001^{***}$ \\
    & Shimmer & \textbf{-0.45 (0.19)} & 0.64 [0.43,0.93] & 0.020* \\
    & Jitter & -0.02 (0.17) & 0.98 [0.70,1.37] & 0.911 \\
    & L(GK) & 0.02 (0.28) & 1.03 [0.59,1.79] & 0.930 \\
    \hline
    \multirow{4}{*}{CN} 
    & G(M) & \textbf{0.74 (0.28)} & \textbf{2.10 [1.23,3.68]} & $0.008^{**}$ \\
    & Speech Portion & \textbf{-1.91 (0.17)} & \textbf{0.15 [0.10,0.20]} & $<0.001^{***}$ \\
    & Shimmer & -0.36 (0.21) & 0.70 [0.47,1.05] & 0.082 \\
    & Jitter & -0.16 (0.20) & 0.85 [0.56,1.26] & 0.434 \\
    \hline
    \multirow{4}{*}{GK} 
    & G(M) & 0.28 (0.21) & 1.32 [0.88,1.99] & 0.184 \\
    & Speech Portion & \textbf{-1.12 (0.13)} & \textbf{0.33 [0.25,0.42]} &  $<0.001^{***}$ \\
    & Jitter & 0.06 (0.15) & 1.06 [0.79,1.43] & 0.688 \\
    & Shimmer & -0.03 (0.14) & 0.97 [0.73,1.28] & 0.827 \\
    \hline
  \end{tabular}}
  \vspace{0.5em}
  
  \footnotesize{Note: Speech Portion, Shimmer, and Jitter have been standardized.}
\end{table}

\subsection{Cross-linguistic Comparison Results}

\begin{figure}[t]
  \centering
  \includegraphics[width=0.88\linewidth]{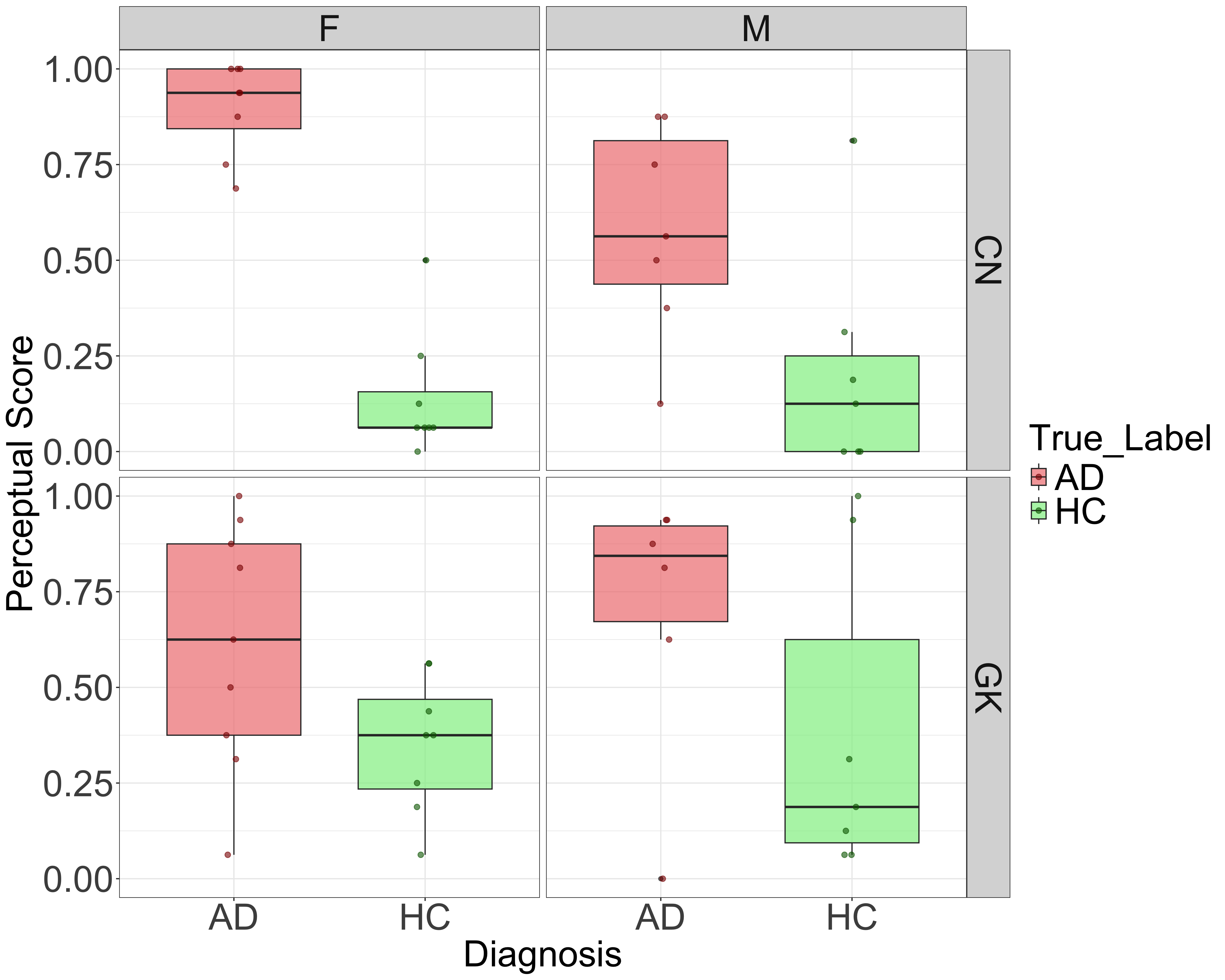}
  \caption{Perceptual Scores Distribution by Language and Gender.}
  \label{fig:crosslingistic}
\end{figure}

This section presents a comparison of the responses from 16 Chinese listeners to CN (30 stimulis) and GK (30 stimulis) speech recordings. Figure~\ref{fig:crosslingistic} presents the distribution of perceptual scores by language and gender.

The Chinese and Greek model results are shown in Table~\ref{tab:vertical}. The GLMER model (Equation~\ref{glmer3}) reveals that male speech was more likely to be perceived as AD in the CN dataset ($p = 0.008$, Odds Ratio = 2.10). In contrast, the proportion of speech had a significant negative effect on AD perception ($p < 0.001$, Odds Ratio = 0.15), with larger speech portions less likely to be perceived as AD. The effects of shimmer and jitter were marginally significant or not significant in this dataset.

\begin{equation}
\begin{split}
   \text{Response} &\sim   \text{Gender} +  \text{Speech\ Portion}  + \text{Jitter} \\ 
  &+ \text{Shimmer} + (1 | \text{Listener\ ID})
  \label{glmer3} 
\end{split}
\end{equation}

\subsubsection{Mandarin Data: Gender's Impact on AD Perception}

In the CN dataset, male speech was more likely to be perceived as AD ($p = 0.008$, Odds Ratio = 2.10). The proportion of speech had a significant negative impact on AD perception ($p < 0.001$, Odds Ratio = 0.15). Shimmer and jitter did not show statistically significant effects.

\subsubsection{Greek Data: Gender’s Influence on AD Perception}

In the GK dataset, male speech showed no significant impact on AD perception ($p = 0.184$), with a slight trend towards male speech being more likely to be perceived as AD. Speech portion had a significant negative impact on AD perception ($p < 0.001$, Odds Ratio = 0.33), while shimmer and jitter did not show significant effects.

\section{Discussion and Conclusion}

In this study, we have identified a significant gender bias in the perception of AD through speech, particularly when evaluating male speech. Our results demonstrate that male speech is more frequently classified as AD compared to female speech, suggesting that gender influences the perception of cognitive decline in speech samples.

We also observed that specific speech features, such as shimmer, play a significant role in AD perception, with higher shimmer values associated with a greater likelihood of AD perception. Additionally, the proportion of speech within the audio clip negatively affected AD perception, with more extended speech segments less likely to be identified as AD. These findings suggest that prosodic features, which are sensitive to the individual’s cognitive state, might be more telling than the sheer amount of speech produced. Interestingly, jitter did not significantly affect AD perception in this study, indicating that speech characteristics are not equally impactful in AD perception. 

While our findings reveal a significant gender bias in speech-based AD perception, it is important to note that the effects of gender on perception were more pronounced in Mandarin Chinese compared to Greek. This cross-linguistic difference might be influenced by listeners' familiarity with language. In the case of Greek, which the listeners were less familiar with, the gender bias in perception was weaker. Further research is needed to understand the factors contributing to this difference.

Our study highlights the importance of addressing gender bias in the development of AD detection models. The reliance on speech-based features in AD detection, while promising, may inadvertently reinforce such biases, leading to misdiagnoses or unequal diagnostic accuracy. Future research should explore ways to mitigate these biases, including the development of more robust and equitable models that can account for gender-related variations in speech without compromising diagnostic performance.

\bibliographystyle{splncs04} 
\bibliography{NCMMSC2025/samplepaper}         
\end{document}